\documentclass[10pt,twocolumn,letterpaper]{article}
\usepackage{cvpr}
\usepackage{times}
\usepackage{epsfig}
\usepackage{graphicx}
\usepackage{subfig}
\usepackage{amsmath}
\usepackage{amssymb}
\usepackage{algorithm}
\usepackage{ifthen}
\usepackage{glossaries}
\usepackage{authblk}
\usepackage{mathtools}
\usepackage[linesnumbered,ruled,vlined,algo2e]{algorithm2e}
\graphicspath{{figures/}{../figures/}}
\usepackage{subfiles}

\graphicspath{{figures/}{../figures/}}

\SetCommentSty{mycommfont}

\SetKwInput{KwInput}{Input}                
\SetKwInput{KwOutput}{Output}              

\newacronym{bn}{BN}{Batch Normalization}
\newacronym{gn}{GN}{Group Normalization}
\newacronym{ln}{LN}{Layer Normalization}
\newacronym{pn}{PN}{Positional Normalization}
\newacronym{in}{IN}{Instance Normalization}
\newacronym{wn}{WN}{Weight Normalization}
\newacronym{lcn}{LCN}{Local Context Normalization}
\newacronym{lrn}{LRN}{Local Response Normalization}
\newacronym{iou}{IoU}{Intersection-over-Union}
\newacronym{mse}{MSE}{mean-squared error}
\newacronym{mlp}{MLP}{multilayer perceptron}
\newacronym{sgd}{SGD}{stochastic gradient descent}
\newacronym{ps}{PS}{Panoptic Segmentation}
\usepackage{url}

\usepackage[pagebackref=true,breaklinks=true,colorlinks,bookmarks=false]{hyperref}

\cvprfinalcopy 

\ifcvprfinal\fi

\begin{document}

\title{Local Context Normalization: Revisiting Local Normalization}

\author[1, 4]{Anthony Ortiz \thanks{Work partially done while author was interning at Microsoft Research}}
\author[3]{Caleb Robinson}
\author[2]{Dan Morris}
\author[1]{Olac Fuentes}
\author[1]{Christopher Kiekintveld}
\author[1]{Md Mahmudulla Hassan}
\author[2]{Nebojsa Jojic \thanks{Correspondence: jojic@microsoft.com, anthonymlortiz@gmail.com}}

\affil[1]{The University of Texas at El Paso}
\affil[2]{Microsoft Research}
\affil[3]{Georgia Institute of Technology}
\affil[4]{Microsoft AI for Good Research Lab}

\maketitle

\begin{abstract}
Normalization layers have been shown to improve convergence in deep neural networks, and even add useful inductive biases.  
In many vision applications the local spatial context of the features is important, but most common normalization schemes including \gls{gn}, \gls{in}, and \gls{ln} normalize over the entire spatial dimension of a feature. This can wash out important signals and degrade performance. For example, in applications that use satellite imagery, input images can be arbitrarily large; consequently, it is nonsensical to normalize over the entire area. \gls{pn}, on the other hand, only normalizes over a single spatial position at a time. A natural compromise is to normalize features by local context, while also taking into account group level information. In this paper, we propose {\em \gls{lcn}}: a normalization layer where every feature is normalized based on a window around it and the filters in its group. 
We propose an algorithmic solution to make \gls{lcn} efficient for arbitrary window sizes, even if every point in the image has a unique window. \gls{lcn} outperforms its \gls{bn}, \gls{gn}, \gls{in}, and \gls{ln} counterparts for object detection, semantic segmentation, and instance segmentation applications in several benchmark datasets, while keeping performance independent of the batch size and facilitating transfer learning.
\end{abstract}

\section{Introduction}

\begin{figure*}[!tbp]
    \centering
        \includegraphics[width=\textwidth]{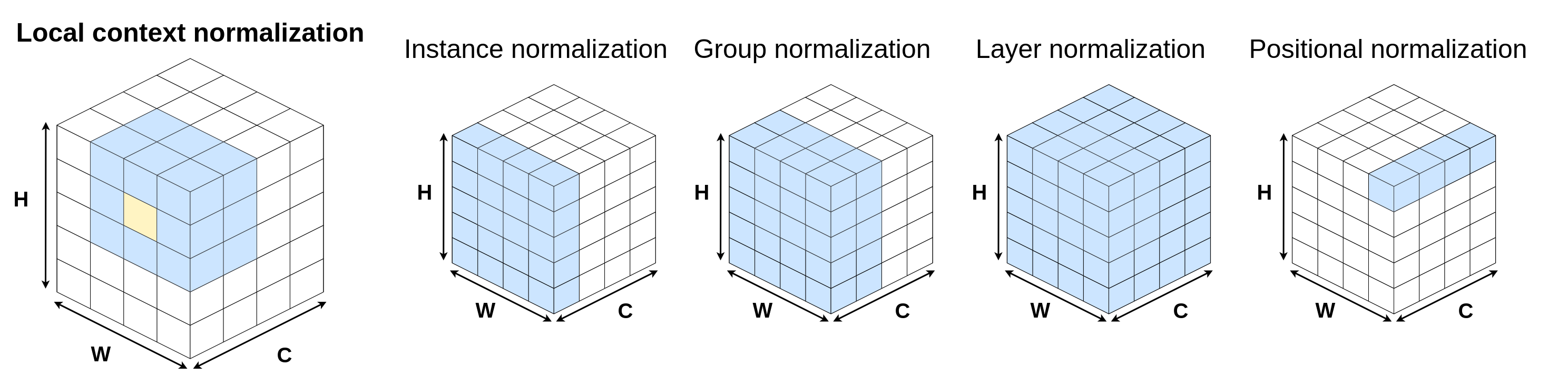}
    \caption{\textbf{Proposed \textit{Local Context Normalization} (LCN) layer}. LCN normalizes each value in a channel according to the values in its feature group and spatial neighborhood. The figure shows how our proposed method compares to other normalization layers in terms of which features are used in normalization (shown in blue), where \textbf{H}, \textbf{W}, and \textbf{C}  are the height, width, and number of channels in the output volume of a convolutional layer.}
    \label{fig:normalization}
\end{figure*}

A variety of neural network normalization layers have been proposed in the literature to aid in convergence and sometimes even add desirable inductive bias. 

Batch Normalization (\gls{bn}) is a subtractive and divisive feature normalization scheme widely used in deep learning architectures~\cite{ioffe2015batch}. Recent research has shown that \gls{bn} facilitates convergence of very deep learning architectures by smoothing the optimization landscape~\cite{santurkar2018does}. \gls{bn} normalizes the features by the mean and variance computed within a mini-batch. Using the batch dimension while calculating the normalization statistics has two main drawbacks:

\begin{itemize}
  \item Small batch sizes affect model performance because the mean and variance estimates are less accurate.
  \item Batches might not exist during inference, so the mean and variance are pre-computed from the training set and used during inference. Therefore, changes in the target data distribution lead to issues while performing transfer learning, since the model assumes the statistics of the original training set~\cite{rebuffi2017learning}.
\end{itemize}

To address both of these issues, Group Normalization (\gls{gn}) was recently proposed by Wu and He~\cite{wu2018group}. \gls{gn} divides channels into groups and normalizes the features by using the statistics within each group. \gls{gn} does not exploit the batch dimension so the computation is independent of batch sizes and model performance does not degrade when the batch size is reduced. \gls{gn} shows competitive performance with respect to \gls{bn} when the batch size is small; consequently, \gls{gn} is being quickly adopted for computer vision tasks like segmentation and video classification, since batch sizes are often restricted for those applications. When the batch size is sufficiently large, \gls{bn} still outperforms \gls{gn}.

\gls{bn}, \gls{gn}, \gls{in}, and \gls{ln} all perform ``global'' normalization where spatial information is not exploited, and all features are normalized by a common mean and variance value. We argue that for the aforementioned applications, local context matters. To incorporate this intuition we propose {\em Local Context Normalization} (\gls{lcn}) as a normalization layer which takes advantage of the context of the data distribution by normalizing each feature based on the statistics of its local neighborhood and corresponding feature group. \gls{lcn} is in fact inspired by computational neuroscience, specifically the {\em contrast normalization} approach leveraged by the human vision system~\cite{lyu2008nonlinear}, as well as early generative modeling approaches to co-segmentation \cite{pim, stel, locus}, where the reasoning about pixel labels is based on shared self-similarity patterns within an image or image window, rather than on shared features across images. \gls{lcn} provides a performance boost over all previously-proposed normalization techniques, while keeping the advantages of being computationally agnostic to the batch size and suitable for transfer learning. We empirically demonstrate the performance benefit of \gls{lcn} for object detection as well as semantic and instance segmentation. 

Another issue with \gls{gn} is that because it performs normalization using the entire spatial dimension of the features, when it is used for inference in applications where input images need to be processed in patches, just shifting the input patch for a few pixels produces different predictions.  This is a common scenario in geospatial analytics and remote sensing applications where the input tends to cover an immense area~\cite{ortiz2018integrated,robinson2019large,ortiz2018defense}. Interactive fine-tuning applications like ~\cite{robinson2019human} become infeasible using \gls{gn}, since a user will not be able to recognize whether changes in the predictions are happening because of fine-tuning or simply because of changes in the image input statistics. With \gls{lcn}, predictions depend only on the statistics within the feature neighborhood; inference does not change when the input is shifted. 

\section{Related Work}

\paragraph{Normalization in Neural Networks.} Since the early days of neural networks, it has been understood that input normalization usually improves convergence~\cite{lecun1998efficient,lecun1998gradient}. LeCun et al. showed that convergence in neural networks is faster if the average of each input variable to any layer is close to zero and their covariances are about the same~\cite{lecun1998gradient}. Many normalization schemes have been proposed in the literature since then~\cite{lyu2008nonlinear,jarrett2009best,krizhevsky2012imagenet,ioffe2015batch, lei2016layer, ulyanov2016instance, wu2018group}. A Local Contrast Normalization Layer was introduced by \cite{jarrett2009best}, later referred to as \gls{lrn}. A modification of this original version of \gls{lrn} was used by the original AlexNet paper which won the Imagenet~\cite{imagenet_cvpr09} challenge in 2012~\cite{krizhevsky2012imagenet}, as well as the 2013 winning entry~\cite{zeiler2014visualizing}. Most popular deep learning architectures until 2015 including Overfeat and GoogLeNet~\cite{sermanet2013overfeat, szegedy2015going} also used \gls{lrn}, which normalizes based on the statistics in a very small window (at most $9\times9$) around each feature. 

 After Ioffe et al. proposed \gls{bn} in 2015, the community moved towards global normalization schemes where the statistics are computed along entire spatial dimensions~\cite{ioffe2015batch}. \gls{bn} normalizes the feature maps of a given mini-batch along the batch dimension. For convolutional layers the mean and variance are computed over both the batch and spatial dimensions, meaning that each location in the feature map is normalized in the same way. Mean and variance are pre-computed on the training set and used at inference time, so when presented with any distribution shift in the input data, \gls{bn} produces inconsistency at the time of transfer or inference~\cite{rebuffi2017learning}. Reducing the batch size also affects \gls{bn} performance as the estimated statistics are less accurate. 

Other normalization methods~\cite{ulyanov2016instance, wu2018group, lei2016layer} have been proposed to avoid exploiting the batch dimension. \gls{ln}~\cite{lei2016layer} performs normalization along the channel dimension, \gls{in}~\cite{ulyanov2016instance} performs normalization for each sample, and \gls{gn} uses the mean and variance from the entire spatial dimension and a group of feature channels. See Figure \ref{fig:normalization} for a visual representation of different normalization schemes. Instead of operating on features, \gls{wn} normalizes the filter weights~\cite{salimans2016weight}. These strategies do not suffer from the issues caused by normalizing along the batch dimension, but they have not been able to approach \gls{bn} performance in most visual recognition applications. Wu and He recently proposed \gls{gn},  which is able to match \gls{bn} performance on some computer vision tasks when the batch size is small~\cite{wu2018group}. All of these approaches perform global normalization, which might wipe out local context. Our proposed \gls{lcn} takes advantages of both local context around the features and improved convergence from global normalization methods.

\paragraph{Contrast Enhancement.} In general, contrast varies widely across a typical image. Contrast enhancement is used to boost contrast in the regions where it is low or moderate, while leaving it unchanged where it is high. This requires that the contrast enhancement be adapted to the local image content. Contrast normalization is inspired by computational neuroscience models~\cite{jarrett2009best, lyu2008nonlinear} and reflects certain aspects of human visual perception. This inspired early normalization schemes for neural networks, but contrast enhancement has not been incorporated into recent normalization methods. Perin et al. showed evidence for synaptic clustering, where small groups of neurons (a few dozen) form small-world networks without hubs~\cite{perin2011synaptic}. For example, in each group, there is an increased probability of connection to other members of the group, not just to a small number of central neurons, facilitating inhibition or excitation within a whole group. Furthermore, these cell assemblies are interlaced so that together they form overlapping groups. Such groups could in fact implement \gls{lcn}. These groups could also implement more extreme color and feature invariance as in probabilistic index map (PIM) models \cite{pim,locus,stel}, which assume that the spatial clustering pattern of features (segmentation) is shared across images but the palette (feature intensities in each cluster) can vary freely. PIMs are naturally suited to co-segmentation applications. LCN also emphasizes local similarities among pixel features, but preserves some intensity information, as well. 

Local contrast enhancement has been applied in computer vision to pre-process input images~\cite{pinto2008real,sermanet2012convolutional} ensuring that contrast is normalized across a very small window ($7\times7$ or $9\times9$ traditionally). Local contrast normalization was essential for the performance of the popular Histogram of Oriented Gradients (HOG) feature descriptors~\cite{dalal2005histograms}. In this work, we propose applying a similar normalization not only at the input layer, but in all layers of a neural network, to groups of neurons.

\section{Local Context Normalization}

\subsection{Formulation}

\paragraph{Local Normalization} In the \gls{lrn} scheme proposed by~\cite{jarrett2009best}, every feature $x_{i,h,w}$ -- where i  refers to channel i and h,w refer to spatial position  of the feature -- is normalized by equation~\ref{eq:LRN}, where $W_{pq}$ is a Gaussian weighting window of size $9\times9$, $\sum_{pq} W_{pq}= 1$, $c$ is set to be $mean(\sigma_{hw})$, and $\sigma_{hw}$ is the weighted standard deviation of all features over a small spatial neighborhood. $h$ and $w$ are spatial coordinates, and $i$ is the feature index.
\begin{equation} 
    \widehat{x}_{ihw} = \frac{x_{ihw} - \sum_{pq} W_{pq}x_{i,h+p,w+q}}{max(c,\sigma_{hw})}
    \label{eq:LRN}
\end{equation}

\paragraph{Global Normalization} Most recent normalization techniques, including \gls{bn}, \gls{ln}, \gls{in}, and \gls{gn}, apply global normalization. In these techniques, features are normalized following equation \ref{eq:normalization}.
\begin{equation}
    \widehat{x}_{i} = \frac{x_i - \mu_i}{\sigma_i}
    \label{eq:normalization}
\end{equation}
For a 2D image, $i = (i_B , i_C , i_H, i_W )$ is a 4D vector indexing the features in $(B, C, H, W)$ order, where $B$ is the batch axis, $C$ is the channel axis, and $H$ and $W$ are the spatial height and width axes. $\mu$ and $\sigma$ are computed as:

\begin{equation}
    \begin{gathered}
        \mu_i = \frac{1}{m}\sum_{k \in S_i} x_k \\ 
        \sigma_i = \sqrt{\frac{1}{m}\sum_{k \in S_i} (x_k - \mu_i)^2 + \epsilon}
    \end{gathered}
\end{equation}

with $\epsilon$ as a small constant. $S_i$ is the set of pixels in which the mean and standard deviation are computed, and $m$ is the size of this set. As shown by~\cite{wu2018group}, most recent types of feature normalization methods mainly differ in how the set $S_i$ is defined. Figure \ref{fig:normalization} shows graphically the corresponding set $S_i$ for different normalization layers.

For \gls{bn}, statistics are computed along ($B,H,W$):
\begin{equation}
    \label{eq:bn_set}
    \text{BN} \implies S_i = \{k | k_C = i_C\}
\end{equation}

For \gls{ln}, normalization is performed per-sample, within each layer. $\mu$ and $\sigma$ are computed along ($C,H,W$):
\begin{equation}
    \label{eq:ln_set}
    \text{LN} \implies \mathcal{S}_i = \{k \vert k_B = i_B\},
\end{equation}
For \gls{in}, normalization is performed per-sample, per-channel. $\mu$ and $\sigma$ are computed along (H,W):
\begin{equation}
    \label{eq:in_set}
    \text{IN} \implies \mathcal{S}_i = \{k \vert k_B = i_B, k_C = i_C\},
\end{equation}

 For \gls{gn}, normalization is performed per-sample, within groups of size $G$ along the channel axis:
\begin{equation}
    \label{eq:gn_set}
    \text{GN} \implies \mathcal{S}_i = \{k \vert k_B = i_B,
    \lfloor\frac{k_C}{C/G}\rfloor = \lfloor\frac{i_C}{C/G}\rfloor\},
\end{equation}

All global normalization schemes (\gls{gn}, \gls{bn}, \gls{ln}, \gls{in}) learn a per-channel linear transformation to compensate for the change in feature amplitude:

\begin{equation}
    \label{film}
    y_{i} = \gamma\widehat{x}_{i} + \beta
\end{equation}

where $\gamma$ and $\beta$ are learned during training.

\paragraph{Local Context Normalization}
In \gls{lcn}, the normalization statistics $\mu$ and $\gamma$ are computed following equation \ref{eq:normalization} using the set $S_i$ defined by \ref{eq:lcn_set}. We propose performing the normalization per-sample, within a window of size $p \times q$, for groups of filters of size predefined by the number of channels per group $(c\_group)$ along the channel axis, as shown in equation \ref{eq:lcn_set}. instead of number of groups $G$ like commonly done for \gls{gn}, we use $(c\_group)$ as hyper-parameter. We consider windows much bigger than the ones used in \gls{lrn} and can compute $\mu$ and $\gamma$ in a computationally efficient manner. The size $p$ and $q$ should be adjusted according to the input size and resolution and can be different for different layers of the network.

\begin{multline}
    \label{eq:lcn_set}
    \text{LCN} \implies \mathcal{S}_i = \{k \vert k_B = i_B,
    \lfloor\frac{k_C}{c\_group}\rfloor = \lfloor\frac{i_C}{c\_group}\rfloor,\\\
    \lfloor\frac{k_H}{p}\rfloor = \lfloor\frac{i_H}{p}\rfloor,\
    \lfloor\frac{k_W}{q}\rfloor = \lfloor\frac{i_W}{q}\rfloor\},
\end{multline}

\paragraph{Relation to Previous Normalization Schemes}
\gls{lcn} allows an efficient generalization of most previously proposed mini-batch-independent normalization layers. Like \gls{gn}, we perform per-group normalization. If the chosen $p$ is greater than or equal to $H$ and the chosen $q$ is greater than or equal to $W$, \gls{lcn} behaves exactly as \gls{gn}, but keeping the number of channels per group fixed throughout the network instead of the number or groups. If in that scenario the number of channels per group (c\_group) is chosen as the total number of channels (c\_group = C), \gls{lcn} becomes \gls{ln}. If the number of channels per group (c\_group) is chosen as 1 (c\_group = 1), \gls{lcn} becomes \gls{in}.

\subsection{Implementation}

\gls{lcn} can be implemented easily in any framework with support for automatic differentiation like PyTorch~\cite{paszke2017automatic} and TensorFlow~\cite{tensorflow2015-whitepaper}. For an efficient calculation of mean and variance, we used the summed area table algorithm, also known in computer vision as the integral image trick~\cite{viola2001rapid}, along with dilated convolutions~\cite{yu2015multi,chen2014semantic}.  Algorithm \ref{pseudocode} shows the pseudo-code for the implementation of \gls{lcn}. We first create two integral images using the input features and the square of the input features. Then, we apply dilated convolution to both integral images with proper dilation (dilation depends on c\_group, p, and q), kernel and stride of one. This provides us the sum and sum of squares tensors for each feature $x_{ihw}$ within the corresponding window and group. From the sums and sum of square tensors we obtain mean and variance tensors needed to normalize the input features. Note that the running time is constant with respect to the window size making \gls{lcn} efficient for arbitrarily large windows~\footnote{A Python implementation of the proposed \gls{lcn} normalization layer using PyTorch can be found at: \url{https://github.com/anthonymlortiz/lcn}}.

\begin{algorithm}[!tbp]
\DontPrintSemicolon
  \KwInput{ $x$: input features of shape [B, C, H, W],\\
            $c\_group$: number of channels per group ,\\
            $window\_size$: spatial window size  as a tuple (p, q),\\
            $\gamma$, $\beta$: scale and shifting parameters to be learned
            }
  \KwOutput{$\{y = LCN_{\gamma, \beta}(x)\}$ }
         $S \leftarrow dilated\_conv(I(x), d, k)  $\  \tcc*{I(x) is integral image of x, dilation d is (c\_group,p,q), kernel k is a tensor with -1 and 1 to substract or add dimension }
        $S_{sq} \leftarrow dilated\_conv(I(x_{sq}), d, k) $\  \tcp*{I($x_{sq}$) is integral image of $x_{sq}$}
        $\mu \leftarrow \frac{S}{n}$ \           \tcp*{Compute Mean $n= c\_group * p * q $} 
        $\sigma^2 \leftarrow \frac{1}{n}(S_{sq} - \frac{S \odot S}{n})$\       \tcp*{compute Variance} 
        $\widehat{x} \leftarrow \frac{x - \mu}{\sqrt{\sigma^2  + \epsilon}}$\                                \tcp*{Normalize activation}
        $y \leftarrow \gamma\widehat{x}+ \beta$\ \tcp*{Apply affine transform}
\caption{LCN pseudo-code}
\label{pseudocode}
\end{algorithm}

\section{Experimental Results}

In this section we evaluate our proposed normalization layer for the tasks of object detection, semantic segmentation, and instance segmentation in several benchmark datasets, and we compare its performance to the best previously known normalization schemes.

\subsection{Semantic Segmentation on Cityscapes}

Semantic segmentation consists of assigning a class label to every pixel in an image. Each pixel is typically labeled with the class of an enclosing object or region. We test for semantic segmentation on the Cityscapes dataset~\cite{cordts2016cityscapes} which contains 5,000 finely-annotated images. The images are divided into 2,975 training, 500 validation, and 1,525 testing images. There are 30 classes, 19 of which are used for evaluation. 

\begin{table*}[!tbp]
    \setlength{\tabcolsep}{0.1cm}
    \caption{Cityscapes Semantic Segmentation Performance
    \label{cityscapes-results}}
    \centering
    \begin{tabular}{| c | c | c | c | c |}
        \hline Method
            & Normalization & mIoU Class (\%) & Pixel Acc. (\%) & Mean Acc. (\%) \\
         
        \hline  HRNetV2 W48 & BN & 76.22  & \textbf{96.39} & 83.73   \\
        \hline  HRNetV2 W48 & GN & 75.08 &  95.84 & 82.70    \\
        \hline  HRNetV2 W48 & LCN (ours) & \textbf{77.49} & 96.14 & \textbf{84.60} \\
        \hline
        \hline  HRNetV2 W18 Small v1 & BN & 71.27 & \textbf{95.36} & 79.49 \\
        \hline  HRNetV2 W18 Small v1 & IN & 69.74 & 94.92 & 77.77  \\
        \hline  HRNetV2 W18 Small v1 & LN & 66.81 & 94.51 & 75.46  \\
        \hline  HRNetV2 W18 Small v1 & GN & 70.31 & 95.03 & 78.99 \\
        \hline  HRNetV2 W18 Small v1 & LCN (ours) & \textbf{71.77} & 95.26 & \textbf{79.72} \\
        \hline  \multicolumn{2}{|c|}{ $\Delta$ GN} & 1.46 & 0.23 & 0.73 \\
        \hline
    \end{tabular}
\end{table*}

\paragraph{Implementation Details.} We train state-of-the-art HRNetV2~\cite{sun2019high} and HRNetV2-W18-Small-v1 networks as baselines~\footnote{We used the official implementation code from: \url{https://github.com/leoxiaobin/deep-high-resolution-net.pytorch}}. We follow the same training protocol as ~\cite{sun2019high}. The data is augmented by random cropping (from 1024 $\times$ 2048 to 512 $\times$ 1024), random scaling in the range of [0.5, 2], and random horizontal flipping. We use the Stochastic Gradient Descent (SGD) optimizer with a base learning rate of 0.01, momentum of 0.9, and weight decay of 0.0005. The poly learning rate policy with the power of 0.9 is used for reducing the learning rate as done in ~\cite{sun2019high}. All the models are trained for 484 epochs. We train HRNetV2 using four GPUs and a batch size of two per GPU. We then substitute sync-batch normalization layers by \gls{bn}, \gls{gn}, \gls{lcn} and compare results. We do exhaustive comparisons using HRNetV2-W18-Small-v1, which is a smaller version of HRNetV2; all training details are kept the same except for the batch size, which is increased to four images per GPU for faster training.   

\paragraph{Quantitative Results.}Table \ref{cityscapes-results} shows the performance of the  different normalization layers on the Cityscapes validation set. In addition to the mean of class-wise intersection over union (mIoU), we also report pixel-wise accuracy (Pixel Acc.) and mean of class-wise pixel accuracy (Mean Acc.). 

\begin{table*}[!tbp]
    \setlength{\tabcolsep}{0.1cm}
    \caption{GN Performance for Different Numbers of Groups
    \label{groups-gn-results}}
    \centering
    \begin{tabular}{ | c | c | c | c | c |}
        \hline Method
            & Number of Groups & mIoU Class (\%) & Pixel Acc. (\%) & Mean Acc. (\%)\\
        \hline  HRNetV2 W18 Small v1  & 1 (=LN) & 66.81 & 94.51 & 75.46  \\
        \hline  HRNetV2 W18 Small v1  & 2 & 69.28 & 94.78 & 77.39 \\
        \hline  HRNetV2 W18 Small v1  & 4 & 67.00 & 94.50 & 76.13  \\
        \hline  HRNetV2 W18 Small v1  & 8 & 67.67 & 94.76 & 75.81\\
        \hline  HRNetV2 W18 Small v1  & 16 & \textbf{70.31} & \textbf{95.03} & \textbf{78.99}\\
        \hline  HRNetV2 W18 Small v1  & C (=IN) & 69.74 & 94.92 & 77.77\\
        \hline
    \end{tabular}
\end{table*}

We observe that our proposed normalization layer outperforms all other normalization techniques including \gls{bn}. \gls{lcn} is almost 1.5\% better than the best \gls{gn} configuration in terms of mIoU. For \gls{lcn}, c\_group was chosen as 2, with a window size of 227 $\times$ 227 ($p$ = $q$ = 227) for HRNetV2 W18 Small v1 and 255 $\times$ 255 for HRNetV2 W48. For \gls{gn}, we tested different numbers of groups as shown in Table \ref{groups-gn-results}, and we report the best (using 16 groups) for comparison with other approaches in Table \ref{cityscapes-results}. Table \ref{groups-gn-results} shows that \gls{gn} is somewhat sensitive to the number of groups, ranging from 67\% to 70.3\% mIoU. Table \ref{groups-gn-results} also shows results for \gls{in} and \gls{ln}, both of which perform worse than the best \gls{gn} performance. These results were obtained using HRNetV2-W18-Small-v1 network architecture. It is important to mention that we used the same learning rate values to train all models, which implies that \gls{lcn} still benefits from the same fast convergence as other global normalization techniques; this is not true for local normalization schemes such as \gls{lrn}, which tend to require lower learning rates for convergence.

\begin{figure*}[ht]\centering
    \centering
        \includegraphics[width=\textwidth]{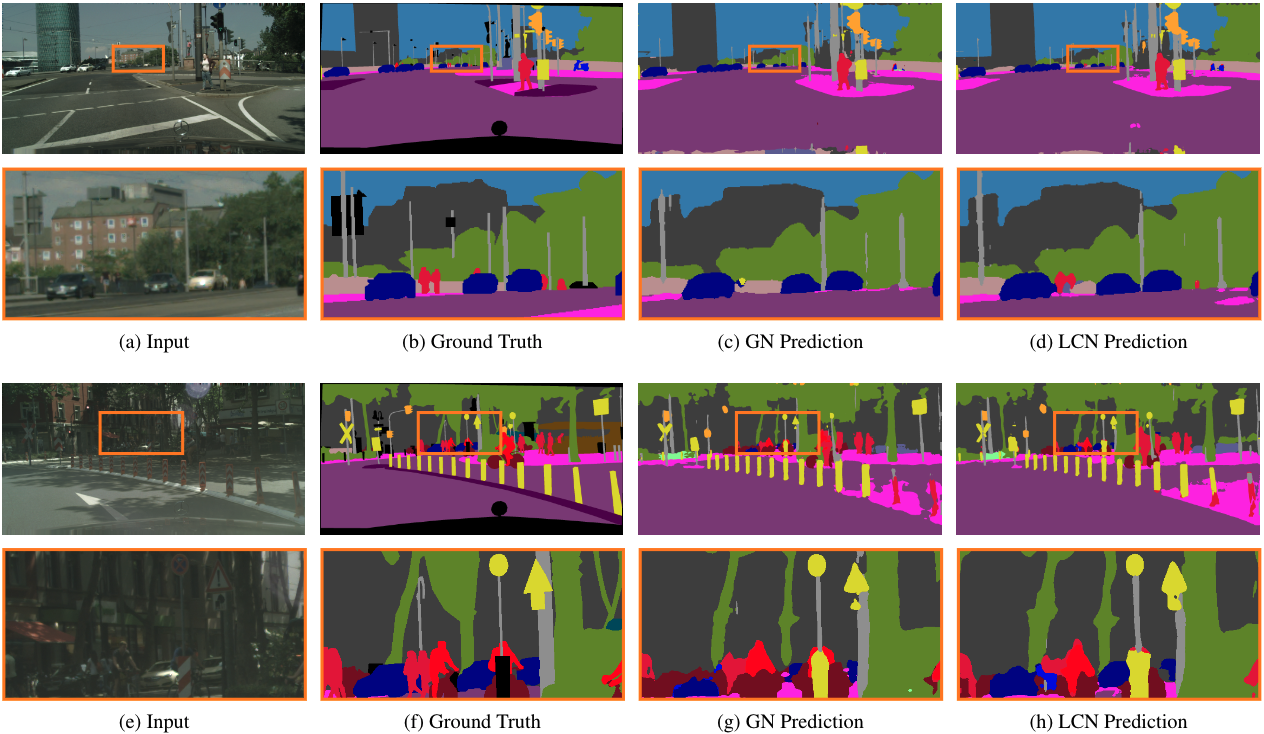}
    \caption{ \textbf{Qualitative results on Cityscapes.} Going from left to right, this figure shows: \textbf{Input}, \textbf{Ground Truth}, \textbf{Group Norm Predictions}, and \textbf{Local Context Norm Predictions}. The second and fourth rows are obtained by maximizing the orange area from the images above. We observe how \gls{lcn} allows the model to detect small objects missed by \gls{gn} and offers sharper and more accurate predictions.}
    \label{fig:qualitative-results}
\end{figure*}

\paragraph{Sensitivity to Number of Channels per Group.} We tested the sensitivity of \gls{lcn} to the number of channels per group (c\_group) parameter by training models for different values of c\_group while keeping the window size fixed to 227 $\times$ 227 ($p$ = $q$ = 227). Table \ref{channels-lcn-results} shows the performance of \gls{lcn} for the different number of channels per group, which is fairly stable among all configurations.

\begin{table*}[h]
    \setlength{\tabcolsep}{0.1cm}
    \caption{LCN sensitivity to number of channels per group for a fixed window size (227, 227)
    \label{channels-lcn-results}}
    \centering
    \begin{tabular}{ | c | c | c | c | c |}
        \hline Method
            & Channels per Group & mIoU Class (\%) & Pixel Acc. (\%) & Mean Acc. (\%)\\
        \hline  HRNetV2 W18 Small v1  & 2 & \textbf{71.77} & \textbf{95.26} & \textbf{79.72} \\
        \hline  HRNetV2 W18 Small v1  & 4 & 70.26 & 95.07 & 78.49 \\
        \hline  HRNetV2 W18 Small v1  & 8 & 70.14 & 94.97 & 78.11 \\
        \hline  HRNetV2 W18 Small v1  & 16 & 70.11 & 94.78 & 79.10\\
        \hline
    \end{tabular}
\end{table*}

\paragraph{Sensitivity to Window Size.} We also tested how \gls{lcn} performance varies with respect to changes in window size while keeping the number of channels per group fixed. The results are shown in Table \ref{sensitivity-ws}. The bigger the window size is the closer \gls{lcn} gets to \gls{gn}. When the window size (p, q) is equal to the entire spatial dimensions \gls{lcn} becomes \gls{gn}. From Table \ref{sensitivity-ws} we see how performance decreases as the window size gets closer to the \gls{gn} equivalent.

\begin{table*}[!tbp]
    \setlength{\tabcolsep}{0.1cm}
    \caption{LCN sensitivity to Window Size
    \label{sensitivity-ws}}
    \centering
    \begin{tabular}{ | c | c | c | c | c |  }
        \hline Method
            & Window Size & mIoU Class (\%) & Pixel Acc. (\%) & Mean Acc. (\%) \\
       
        \hline  HRNetV2 Small v1 & 199 & 71.55 & 95.18 & \textbf{79.89} \\
        \hline  HRNetV2 Small v1 & 227 & 71.77 & \textbf{95.26} & 79.72  \\
        \hline  HRNetV2 Small v1 & 255 & \textbf{71.80} & 95.18 & 79.26   \\
        \hline  HRNetV2 Small v1 & 383 & 70.09 & 95.06 & 77.64 \\
        \hline  HRNetV2 Small v1 & 511 & 70.03 & 95.09 & 77.94\\
        \hline  HRNetV2 Small v1 & all/GN & 70.30 & 95.04 & 78.97\\
        \hline
    \end{tabular}
\end{table*}

\paragraph{Qualitative Results}

Figure \ref{fig:qualitative-results} shows two randomly selected examples of the semantic segmentation results obtained from HRNetV2-W18-Small-v1 using \gls{gn} (last column) and \gls{lcn} (second-to-last column) as the normalization layers. The second and fourth rows are obtained by maximizing the orange area from the images above them. By zooming in and looking at the details in the segmentation results, we see that \gls{lcn} allows sharper and more accurate predictions. Carefully looking at the second row, we can observe how using \gls{gn} HRNet misses pedestrians, which are recognized when using \gls{lcn}. From the last row, we can see that using \gls{lcn} results in sharper and less discontinuous predictions. \gls{lcn} allows HRNet to distinguish between the bike and the legs of the cyclist while \gls{gn} cannot. \gls{lcn} also provides more precise boundaries for the cars in the background than \gls{gn}. 

\subsection{Object Detection and Instance Segmentation on Microsoft COCO Dataset}

\begin{table*}[!tbp]
        \setlength{\tabcolsep}{0.2cm}
            \caption{Detection and Instance Segmentation Performance on the Microsoft Coco Dataset
            \label{coco-results}}
            \centering
            \begin{tabular}{ c | c | c | c | c | c | c   }
                \hline Method
                    & AP$^{bbox}$ (\%)& AP$_{50}^{bbox}$ (\%)& AP$_{75}^{bbox}$ (\%)& AP$^{mask}$ (\%)& AP$_{50}^{mask}$ (\%)& AP$_{75}^{mask}$\\
                \hline R50 BN & 37.47 & 59.15 & 40.76 & 34.06 & 55.74 & 36.04 \\
                \hline R50 GN & 37.34 & 59.65 & 40.34 & 34.33 & 56.53 & 36.31 \\
                \hline R50 LCN (Ours) & \textbf{37.90} & \textbf{59.82} & \textbf{41.16} & \textbf{34.50} & \textbf{56.81} & \textbf{36.43} \\
                \hline
            \end{tabular}
        \end{table*}
        
We evaluate our \gls{lcn} against previously-proposed normalization schemes for object detection and instance segmentation. Object detection involves detecting instances of objects from a particular class in an image. Instance segmentation involves detecting and segmenting each object in an image. The Microsoft COCO dataset~\cite{lin2014microsoft} is a high-quality dataset which provides labels appropriate for both detection and instance segmentation and is the standard dataset for both tasks. The annotations include both pixel-level segmentation masks and bounding boxes for  objects belonging to 80 categories.
        
 These computer vision tasks in general benefit from higher-resolution input. We experiment with the Mask R-CNN baselines~\cite{he2017mask}, implemented in the publicly available Detectron codebase. We replace \gls{bn} and/or \gls{gn} by \gls{lcn} during finetuning, using the model pre-trained from ImageNet using GN. We fine-tune with a batch size of one image per GPU and train the model using four GPUs.

The models are trained in the COCO~\cite{lin2014microsoft} train2017 set and evaluated in the COCO val2017 set (a.k.a. minival). We report the standard COCO metrics of Average Precision (AP), $AP_{50}$, and $AP_{75}$, for both bounding box detection ($AP^{bbox}$) and instance segmentation ($AP^{mask}$).

Table \ref{coco-results} shows the performance of the different normalization techniques\footnote{Our results differ slightly from the ones reported in the original paper, but this should not affect the comparison across normalization schemes.}. \gls{lcn} outperforms both \gls{gn} and \gls{bn} by a substantial margin in all experiments, even using hyper-parameters tuned for the other schemes. 

\begin{table*}[!tbp]
    \setlength{\tabcolsep}{0.2cm}
    \caption{Image Classification Error on Imagenet
    \label{imagenet-results}}
    \centering
    \begin{tabular}{| c | c | c | c |    }
        \hline Network Architecture
            & Normalization & Top 1 Err. (\%) & Top 5 Err. (\%) \\
        \hline Resnet 50 & BN & 23.59 &  6.82  \\
        \hline Resnet 50 & GN & 24.24 &  7.35  \\
        \hline Resnet 50 & LCN & 24.23 & 7.22  \\
        \hline
    \end{tabular}
\end{table*}

\subsection{Image Classification in ImageNet}

We also experiment with image classification using the ImageNet dataset~\cite{imagenet_cvpr09}. In this experiment, images must be classified into one of 1000 classes. We train on all training images and evaluate on the 50,000 validation images, using the ResNet models~\cite{he2016deep}.

\paragraph{Implementation Details.} As in most reported results, we use eight GPUs to train all models, and the batch mean and variance of \gls{bn} are computed within each GPU. We use He's initialization~\cite{he2015delving} to initialize  convolution weights.  We train all models for 100 epochs, and decrease the learning rate by $10\times$ at 30, 60, and 90 epochs.

During training, we adopt the data augmentation of Szegedy et al.~\cite{szegedy2015going} as used in~\cite{wu2018group}. We evaluate the top-1 classification error on the center crops of $224 \times 224$ pixels in the validation set. To reduce random variations, we report the median error rate of the final five epochs~\cite{goyal2017accurate}. 

As in~\cite{wu2018group} our baseline is the ResNet trained with \gls{bn}~\cite{he2016deep}. To compare with \gls{gn} and \gls{lcn}, we replace \gls{bn} with the specific variant. We use the same hyper-parameters for all models. We set the number of channels per group for \gls{lcn} as 32, and we used $p = q = 127$ for the window size parameters. Table \ref{imagenet-results} shows that \gls{lcn} offers similar performance as \gls{gn}, but we don't see the same boost in performance observed for object detection and image segmentation. We hypothesize that this happens because image classification is a global task which might not benefit from local context.

\begin{table*}[!tbp]
    \setlength{\tabcolsep}{0.1cm}
    \caption{Performance in INRIA Aerial Image Labeling Dataset. \gls{lcn} outperforms all the other normalization layers overall.
    \label{inria-results}}
    \centering
    \begin{tabular}{ c | c | c | c | c | c | c | c | c | c | c | c | c  }
        \hline Method &
            \multicolumn{2}{|c}{Bellingham} &
            \multicolumn{2}{|c}{Bloomington} &
            \multicolumn{2}{|c}{Innsbruck} &
            \multicolumn{2}{|c}{San Francisco} &
            \multicolumn{2}{|c}{East Tyrol} &
            \multicolumn{2}{|c}{\textbf{Overall}} \\
        \hline & IoU & Acc. & IoU & Acc. & IoU & Acc. & IoU & Acc. & IoU & Acc. & IoU & Acc.\\
        \hline U-Net + BN & \textbf{65.37} & \textbf{96.53} & 55.07 & 95.83 &       67.62 & 96.08 & 72.80 & 91.00 & 67.00 & 96.91 & 67.98 & 95.27\\
        \hline U-Net + GN & 55.48 & 93.38 & 55.47 & 94.41 & 58.93 & 93.77 &         72.12 & 89.56 & 62.27 & 95.73 & 63.71 & 93.45\\
        \hline U-Net + LCN & 63.61 & 96.26 & \textbf{60.47} & \textbf{96.22} &     \textbf{68.99} & \textbf{96.28} & \textbf{75.01} & \textbf{91.46} &     \textbf{68.90} & \textbf{97.19} & \textbf{69.90} & \textbf{95.48}\\
        \hline
        \end{tabular}
\end{table*}

\subsection{Systematic Generalization on INRIA Aerial Imagery Dataset}

The INRIA Aerial Image Labeling Dataset was introduced to test generalization of remote-sensing segmentation models~\cite{maggiori2017dataset}. It includes imagery from 10 dissimilar urban areas in North America and Europe. Instead of splitting adjacent portions of the same images into training and test sets, the splitting was done city-wise. All tiles of five cities were included in the training set and the remaining ones are used as the test set. The imagery is orthorectified~\cite{maggiori2017dataset} and has a spatial resolution of 0.3m per pixel. The dataset covers 810 $km^{2}$ (405 $km^{2}$ for training and 405 $km^{2}$ for the test set). Images were labeled for the semantic classes of building and non-building. 

\paragraph{Implementation Details.} We trained different versions of U-Net~\cite{ronneberger2015u} where just the normalization layer was changed. We trained all models in this set of experiments using $572 \times 572$ randomly sampled patches from all training image tiles.  We used the Adam optimizer with a batch size of 12. All networks were trained from scratch with a starting learning rate of 0.001. We keep the same learning rate for the first 60 epochs and decay it to 0.0001 over the next 40 epochs. Every network was trained for 100 epochs. In every epoch 8,000 patches are seen. Binary cross-entropy loss was used as the loss function.

Table \ref{inria-results} summarizes the performance of the different normalization layers in the INRIA aerial image labeling dataset. Our proposed \gls{lcn} outperforms all the other normalization layers with an overall mIoU almost 2\% higher than the next-best normalization scheme, and more than 6\% better than \gls{gn} in terms of overall IoU. \gls{lcn} provides much better performance than other methods in almost every test city. \gls{lcn} was trained using a $91 \times 91$ window size and four channels per group.

\subsection{Land Cover Mapping}

\begin{table}[h]
    \setlength{\tabcolsep}{0.2cm}
    \caption{Landcover Mapping Tested on Maryland 2013 Test Tiles
    \label{lcm-results}}
    \centering
    \begin{tabular}{| c | c | c |} 
        \hline Method & mIoU (\%) & Pixel Acc. (\%) \\
        \hline UNet + BN & 76.69 & 87.15 \\
        \hline UNet + GN & 74.15 & 85.18 \\
        \hline UNet + LCN & 76.51 & 86.96 \\
        \hline
    \end{tabular}
\end{table} 

Finally, we evaluate \gls{lcn} on a land cover mapping task previously studied in~\cite{robinson2019large,landCoverDatasetOnLila}. Land cover mapping is a semantic image segmentation task where each pixel in an aerial or satellite image must be classified as belonging to one of a variety of land cover classes. 
This process of turning raw remotely sensed imagery into a summarized data product is an important first step in many downstream sustainability related applications. For example, the Chesapeake Bay Conservancy uses land cover data in a variety of settings including determining where to target planting riparian forest buffers~\cite{chesapeakeData}.
The dataset can be found at~\cite{landCoverDatasetOnLila} and contains 4-channel (red, green, blue, and near-infrared), 1m resolution imagery from the National Agricultural Imagery Program (NAIP) and dense pixel labels from the Chesapeake Conservancy's land cover mapping program over 100,000 square miles intersecting 6 states in the northeastern US.
We use the Maryland 2013 subset - training on the 50,000 multi-spectral image patches, each of size $256 \times 256 \times 4$, from the train split. We test over the 20 test tiles\footnote{Consisting of $\sim 900,000,000$ pixels}.  Each pixel must be classified as: water, tree canopy / forest, low vegetation / field, or impervious surfaces.

\paragraph{Implementation Details} We trained different versions of U-Net architecture used on~\cite{robinson2019large} for different normalization layers without doing any data augmentation and compared results. We used the Adam optimizer with a batch size of 96. All networks were trained from scratch for 100 epochs with a starting learning rate of 0.001 with decay to 0.0001 after 60 epochs. The multi-class cross-entropy loss was used as criterion. The best \gls{gn} results are obtained using 8 groups.  \gls{lcn} results are obtained using 4 channels per group and and a $31 \times 31$ window.

Table \ref{lcm-results} shows the mean IoU and Pixel Accuracy of the different normalization layers for land cover mapping. \gls{lcn} outperforms \gls{gn} for this task with performance slightly lower than \gls{bn}. We notice that \gls{lcn} benefits from larger input images. When input images are small like in this setting the performance boost from using \gls{lcn} becomes smaller.

\section{Discussion and Conclusion}

We proposed {\em Local Context Normalization} (\gls{lcn}) as a normalization layer where every feature is normalized based on a window around it and the filters in its group. We empirically showed that \gls{lcn} outperforms all previously-proposed normalization layers for object detection, semantic segmentation, and instance image segmentation across a variety of datasets. The performance of \gls{lcn} is invariant to batch size, and it is well-suited for transfer learning and interactive systems. 

We note that we used hyper-parameters which were already highly optimized for \gls{bn} and/or \gls{gn} without tuning, so it is likely that we could obtain better results with \gls{lcn} by just searching for better hyper-parameters. In our experiments we also do not consider varying the window size for different layers in the network, but it is a direction worth exploring: adjusting the window size during training via gradient descent may further improve performance for \gls{lcn}.  

\section*{Acknowledgement}
The  authors  thank  Lucas  Joppa  and  the  Microsoft  AI  for Earth initiative for their support. A.O. was supported by the Army Research Office under award W911NF-17-1-0370. We thank Nvidia Corporation for the donation of two Titan Xp GPUs used for this research.  
{\small
\bibliographystyle{ieee_fullname}
\bibliography{egbib}

\begin{thebibliography}{10}\itemsep=-1pt

\bibitem{landCoverDatasetOnLila}
Chesapeake land cover.
\newblock Maryland split.

\bibitem{tensorflow2015-whitepaper}
Mart\'{\i}n Abadi, Ashish Agarwal, Paul Barham, Eugene Brevdo, Zhifeng Chen,
  Craig Citro, Greg~S. Corrado, Andy Davis, Jeffrey Dean, Matthieu Devin,
  Sanjay Ghemawat, Ian Goodfellow, Andrew Harp, Geoffrey Irving, Michael Isard,
  Yangqing Jia, Rafal Jozefowicz, Lukasz Kaiser, Manjunath Kudlur, Josh
  Levenberg, Dan Man\'{e}, Rajat Monga, Sherry Moore, Derek Murray, Chris Olah,
  Mike Schuster, Jonathon Shlens, Benoit Steiner, Ilya Sutskever, Kunal Talwar,
  Paul Tucker, Vincent Vanhoucke, Vijay Vasudevan, Fernanda Vi\'{e}gas, Oriol
  Vinyals, Pete Warden, Martin Wattenberg, Martin Wicke, Yuan Yu, and Xiaoqiang
  Zheng.
\newblock {TensorFlow}: Large-scale machine learning on heterogeneous systems,
  2015.
\newblock Software available from tensorflow.org.

\bibitem{chen2014semantic}
Liang-Chieh Chen, George Papandreou, Iasonas Kokkinos, Kevin Murphy, and Alan~L
  Yuille.
\newblock Semantic image segmentation with deep convolutional nets and fully
  connected {CRFs}.
\newblock {\em arXiv preprint arXiv:1412.7062}, 2014.

\bibitem{chesapeakeData}
{Chesapeake Conservancy}.
\newblock Land cover data project 2013/2014.
\newblock
  \url{https://chesapeakeconservancy.org/conservation-innovation-center/high-resolution-data/land-cover-data-project/},
  2016.

\bibitem{cordts2016cityscapes}
Marius Cordts, Mohamed Omran, Sebastian Ramos, Timo Rehfeld, Markus Enzweiler,
  Rodrigo Benenson, Uwe Franke, Stefan Roth, and Bernt Schiele.
\newblock The cityscapes dataset for semantic urban scene understanding.
\newblock In {\em Proceedings of the IEEE conference on computer vision and
  pattern recognition}, pages 3213--3223, 2016.

\bibitem{dalal2005histograms}
Navneet Dalal and Bill Triggs.
\newblock Histograms of oriented gradients for human detection.
\newblock In {\em 2005 IEEE conference on computer vision and pattern
  recognition}, 2005.

\bibitem{imagenet_cvpr09}
J. Deng, W. Dong, R. Socher, L.-J. Li, K. Li, and L. Fei-Fei.
\newblock {ImageNet: A Large-Scale Hierarchical Image Database}.
\newblock In {\em CVPR09}, 2009.

\bibitem{goyal2017accurate}
Priya Goyal, Piotr Doll{\'a}r, Ross Girshick, Pieter Noordhuis, Lukasz
  Wesolowski, Aapo Kyrola, Andrew Tulloch, Yangqing Jia, and Kaiming He.
\newblock Accurate, large minibatch {SGD}: Training {Imagenet} in 1 hour.
\newblock {\em arXiv preprint arXiv:1706.02677}, 2017.

\bibitem{he2017mask}
Kaiming He, Georgia Gkioxari, Piotr Doll{\'a}r, and Ross Girshick.
\newblock Mask {R-CNN}.
\newblock In {\em Proceedings of the IEEE international conference on computer
  vision}, pages 2961--2969, 2017.

\bibitem{he2015delving}
Kaiming He, Xiangyu Zhang, Shaoqing Ren, and Jian Sun.
\newblock Delving deep into rectifiers: Surpassing human-level performance on
  imagenet classification.
\newblock In {\em Proceedings of the IEEE international conference on computer
  vision}, pages 1026--1034, 2015.

\bibitem{he2016deep}
Kaiming He, Xiangyu Zhang, Shaoqing Ren, and Jian Sun.
\newblock Deep residual learning for image recognition.
\newblock In {\em Proceedings of the IEEE Conference on Computer Cision and
  Pattern Recognition}, pages 770--778, 2016.

\bibitem{ioffe2015batch}
Sergey Ioffe and Christian Szegedy.
\newblock Batch normalization: Accelerating deep network training by reducing
  internal covariate shift.
\newblock {\em Proceedings of the International Conference in Machine Learning
  (ICML)}, 2015.

\bibitem{jarrett2009best}
Kevin Jarrett, Koray Kavukcuoglu, Yann LeCun, et~al.
\newblock What is the best multi-stage architecture for object recognition?
\newblock In {\em 2009 IEEE 12th International Conference on Computer Vision},
  pages 2146--2153. IEEE, 2009.

\bibitem{pim}
Nebojsa Jojic and Yaron Caspi.
\newblock Capturing image structure with probabilistic index maps.
\newblock In {\em Proceedings of the 2004 IEEE Computer Society Conference on
  Computer Vision and Pattern Recognition, 2004. CVPR 2004.}, volume~1, pages
  I--I. IEEE, 2004.

\bibitem{stel}
Nebojsa Jojic, Alessandro Perina, Marco Cristani, Vittorio Murino, and Brendan
  Frey.
\newblock Stel component analysis: Modeling spatial correlations in image class
  structure.
\newblock In {\em 2009 IEEE conference on computer vision and pattern
  recognition}, pages 2044--2051. IEEE, 2009.

\bibitem{krizhevsky2012imagenet}
Alex Krizhevsky, Ilya Sutskever, and Geoffrey~E Hinton.
\newblock Imagenet classification with deep convolutional neural networks.
\newblock In {\em Advances in neural information processing systems}, pages
  1097--1105, 2012.

\bibitem{lecun1998gradient}
Yann LeCun, L{\'e}on Bottou, Yoshua Bengio, Patrick Haffner, et~al.
\newblock Gradient-based learning applied to document recognition.
\newblock {\em Proceedings of the IEEE}, 86(11):2278--2324, 1998.

\bibitem{lecun1998efficient}
Yann~A LeCun, L{\'e}on Bottou, Genevieve~B Orr, and Klaus-Robert M{\"u}ller.
\newblock Efficient backprop.
\newblock In {\em Neural networks: Tricks of the trade}, pages 9--48. Springer,
  1998.

\bibitem{lei2016layer}
Jimmy Lei~Ba, Jamie~Ryan Kiros, and Geoffrey~E Hinton.
\newblock Layer normalization.
\newblock {\em arXiv preprint arXiv:1607.06450}, 2016.

\bibitem{lin2014microsoft}
Tsung-Yi Lin, Michael Maire, Serge Belongie, James Hays, Pietro Perona, Deva
  Ramanan, Piotr Doll{\'a}r, and C~Lawrence Zitnick.
\newblock Microsoft {COCO}: Common objects in context.
\newblock In {\em European conference on computer vision}, pages 740--755.
  Springer, 2014.

\bibitem{lyu2008nonlinear}
Siwei Lyu and Eero~P Simoncelli.
\newblock Nonlinear image representation using divisive normalization.
\newblock In {\em 2008 IEEE Conference on Computer Vision and Pattern
  Recognition}, pages 1--8. IEEE, 2008.

\bibitem{maggiori2017dataset}
Emmanuel Maggiori, Yuliya Tarabalka, Guillaume Charpiat, and Pierre Alliez.
\newblock Can semantic labeling methods generalize to any city? the inria
  aerial image labeling benchmark.
\newblock In {\em IEEE International Geoscience and Remote Sensing Symposium
  (IGARSS)}. IEEE, 2017.

\bibitem{ortiz2018defense}
Anthony Ortiz, Olac Fuentes, Dalton Rosario, and Christopher Kiekintveld.
\newblock On the defense against adversarial examples beyond the visible
  spectrum.
\newblock In {\em MILCOM 2018-2018 IEEE Military Communications Conference
  (MILCOM)}, pages 1--5. IEEE, 2018.

\bibitem{ortiz2018integrated}
Anthony Ortiz, Alonso Granados, Olac Fuentes, Christopher Kiekintveld, Dalton
  Rosario, and Zachary Bell.
\newblock Integrated learning and feature selection for deep neural networks in
  multispectral images.
\newblock In {\em Proceedings of the IEEE Conference on Computer Vision and
  Pattern Recognition Workshops}, pages 1196--1205, 2018.

\bibitem{paszke2017automatic}
Adam Paszke, Sam Gross, Soumith Chintala, Gregory Chanan, Edward Yang, Zachary
  DeVito, Zeming Lin, Alban Desmaison, Luca Antiga, and Adam Lerer.
\newblock Automatic differentiation in {PyTorch}.
\newblock In {\em NIPS-W}, 2017.

\bibitem{perin2011synaptic}
Rodrigo Perin, Thomas~K Berger, and Henry Markram.
\newblock A synaptic organizing principle for cortical neuronal groups.
\newblock {\em Proceedings of the National Academy of Sciences},
  108(13):5419--5424, 2011.

\bibitem{pinto2008real}
Nicolas Pinto, David~D Cox, and James~J DiCarlo.
\newblock Why is real-world visual object recognition hard?
\newblock {\em PLoS computational biology}, 4(1):e27, 2008.

\bibitem{rebuffi2017learning}
Sylvestre-Alvise Rebuffi, Hakan Bilen, and Andrea Vedaldi.
\newblock Learning multiple visual domains with residual adapters.
\newblock In {\em Advances in Neural Information Processing Systems}, pages
  506--516, 2017.

\bibitem{robinson2019large}
Caleb Robinson, Le Hou, Kolya Malkin, Rachel Soobitsky, Jacob Czawlytko, Bistra
  Dilkina, and Nebojsa Jojic.
\newblock Large scale high-resolution land cover mapping with multi-resolution
  data.
\newblock In {\em Proceedings of the IEEE Conference on Computer Vision and
  Pattern Recognition}, pages 12726--12735, 2019.

\bibitem{robinson2019human}
Caleb Robinson, Anthony Ortiz, Kolya Malkin, Blake Elias, Andi Peng, Dan
  Morris, Bistra Dilkina, and Nebojsa Jojic.
\newblock Human-machine collaboration for fast land cover mapping.
\newblock {\em AAAI Conference on Artificial Intelligence (AAAI 2020)}, 2020.

\bibitem{ronneberger2015u}
Olaf Ronneberger, Philipp Fischer, and Thomas Brox.
\newblock U-net: Convolutional networks for biomedical image segmentation.
\newblock In {\em International Conference on Medical image computing and
  computer-assisted intervention}, pages 234--241. Springer, 2015.

\bibitem{salimans2016weight}
Tim Salimans and Durk~P Kingma.
\newblock Weight normalization: A simple reparameterization to accelerate
  training of deep neural networks.
\newblock In {\em Advances in Neural Information Processing Systems}, pages
  901--909, 2016.

\bibitem{santurkar2018does}
Shibani Santurkar, Dimitris Tsipras, Andrew Ilyas, and Aleksander Madry.
\newblock How does batch normalization help optimization?
\newblock In {\em Advances in Neural Information Processing Systems}, pages
  2483--2493, 2018.

\bibitem{sermanet2012convolutional}
Pierre Sermanet, Soumith Chintala, and Yann LeCun.
\newblock Convolutional neural networks applied to house numbers digit
  classification.
\newblock In {\em 2012 21st International Conference on Pattern Recognition
  (ICPR 2012)}, pages 3288--3291. IEEE, 2012.

\bibitem{sermanet2013overfeat}
Pierre Sermanet, David Eigen, Xiang Zhang, Micha{\"e}l Mathieu, Rob Fergus, and
  Yann LeCun.
\newblock Overfeat: Integrated recognition, localization and detection using
  convolutional networks.
\newblock {\em arXiv preprint arXiv:1312.6229}, 2013.

\bibitem{sun2019high}
Ke Sun, Yang Zhao, Borui Jiang, Tianheng Cheng, Bin Xiao, Dong Liu, Yadong Mu,
  Xinggang Wang, Wenyu Liu, and Jingdong Wang.
\newblock High-resolution representations for labeling pixels and regions.
\newblock {\em arXiv preprint arXiv:1904.04514}, 2019.

\bibitem{szegedy2015going}
Christian Szegedy, Wei Liu, Yangqing Jia, Pierre Sermanet, Scott Reed, Dragomir
  Anguelov, Dumitru Erhan, Vincent Vanhoucke, and Andrew Rabinovich.
\newblock Going deeper with convolutions.
\newblock In {\em Proceedings of the IEEE conference on computer vision and
  pattern recognition}, pages 1--9, 2015.

\bibitem{ulyanov2016instance}
Dmitry Ulyanov, Andrea Vedaldi, and Victor Lempitsky.
\newblock Instance normalization: The missing ingredient for fast stylization.
\newblock {\em arXiv preprint arXiv:1607.08022}, 2016.

\bibitem{viola2001rapid}
Paul Viola, Michael Jones, et~al.
\newblock Rapid object detection using a boosted cascade of simple features.
\newblock 2001.

\bibitem{locus}
John Winn and Nebojsa Jojic.
\newblock Locus: Learning object classes with unsupervised segmentation.
\newblock In {\em Tenth IEEE International Conference on Computer Vision
  (ICCV'05) Volume 1}, volume~1, pages 756--763. IEEE, 2005.

\bibitem{wu2018group}
Yuxin Wu and Kaiming He.
\newblock Group normalization.
\newblock In {\em European Conference on Computer Vision}, pages 3--19.
  Springer, 2018.

\bibitem{yu2015multi}
Fisher Yu and Vladlen Koltun.
\newblock Multi-scale context aggregation by dilated convolutions.
\newblock {\em arXiv preprint arXiv:1511.07122}, 2015.

\bibitem{zeiler2014visualizing}
Matthew~D Zeiler and Rob Fergus.
\newblock Visualizing and understanding convolutional networks.
\newblock In {\em European conference on computer vision}, pages 818--833.
  Springer, 2014.

\end{thebibliography}
}

\end{document}